\documentclass[journal]{IEEEtran}
\usepackage{ltexpprt}
\usepackage{graphicx}
\usepackage{amssymb}
\usepackage{amsmath}
\usepackage{epstopdf}
\usepackage[tight,footnotesize]{subfigure}
\usepackage{stfloats}
\usepackage{microtype}
\usepackage{units}
\usepackage{algorithm}
\usepackage[noend]{algpseudocode}
\usepackage{url}
\usepackage{bm}
\usepackage{multirow}

\usepackage{float}
\usepackage{adjustbox}

\usepackage{verbatim}

\usepackage{booktabs} 
\usepackage{verbatim} 

\begin{document}
\title{Detecting Electric Devices in 3D Images of Bags}

\author{ Anthony Bagnall, Paul Southam, James Large and Richard Harvey
\thanks{School of Computing Sciences, University of East Anglia, Norwich, NR4 7TJ, UK.}}


\maketitle              

\begin{abstract}
The aviation and transport security industries face the challenge of screening high volumes of baggage for threats and contraband in the minimum time possible. Automation and semi-automation of this procedure offers the potential to increase security by detecting more threats and improve the customer experience by speeding up the process. Traditional 2D x-ray images are often extremely difficult to examine due to the fact that they are tightly packed and contain a wide variety of cluttered and occluded objects.  Because of these limitations, major airports are introducing 3D x-ray Computed Tomography (CT) baggage scanning. We investigate whether we can automate the process of detecting electric devices in these 3D images of luggage. Detecting electrical devices is of particular concern as they can be used to conceal explosives. Given the massive volume of luggage that needs to be screened for this threat, the best way to automate the detection is to first filter whether a bag contains an electric device or not, and if it does, to identify the number of devices and their location. We present an algorithm, Unpack, Predict, eXtract, Repack (UXPR), which involves unpacking through segmenting the data at a range of scales using an algorithm known as the Sieve, predicting whether a segment is electrical or not based on the histogram of voxel intensities, then repacking the bag by ensembling the segments and predictions to identify the devices in bags. Through a range of experiments using data provided by ALERT (Awareness and Localization of Explosives-Related Threats) we show that this system can find a high proportion of devices with unsupervised segmentation if a similar device has been seen before, and shows promising results for detecting devices not seen at all based on the properties of its constituent parts.

\end{abstract}

\section{Introduction}

Baggage screening plays a major role within the aviation and transport security domain. Screening personnel perform the difficult task of examining thousands of bags for wide variety of contraband items. Staff often only have a few seconds to visually interpret x-ray images and determine if there a threat or contraband present. These x-ray images are often extremely difficult to examine due to the fact that they are tightly packed and contain a wide variety of cluttered and occluded objects. With the increase in global travel this task becomes more important and it is increasingly challenging to process baggage in the required time scales. As such there has been an increase in demand for automated and semi-automated threat detection systems that can assist screening personnel. Airport security represents a significant cost to operators and a major inconvenience to passengers. Any improvement in the speed and efficiency of screening, however small, would be hugely beneficial.

One source of improvement is more advanced technology. 3D x-ray Computed Tomography (CT) baggage scanning aims to address some of the limitations in conventional 2D x-ray scanning such as object occlusion, density, clutter and confusion. These scanners generate a series of image slices through the bag, which then can be reconstructed to form a CT 3D volume. The process is similar to medical CT scanners but due to the demand for faster scanning speeds, the resulting volumes have lower voxel resolutions with anisotropic voxels. A number of existing and well known 3D object detection and classification techniques have been applied to these volumes ~\cite{Mouton2015areview}. Our interest lies in detecting electrical devices in 3D CT scans.

Detecting electrical devices is of particular concern as they can be used to conceal explosives. Given the massive volume of luggage that needs to be screened for this threat, the best way to automate the detection is to first filter whether a bag contains an electric device or not, and if it does, to identify the number of devices and their location. We present an algorithm that can segment a bag, classify each segment as electrical or not then amalgamate these predictions to identify the locations of items of interest. Our approach to this problem is based on the basic premise that the scale of 3D images and the lack of annotated training data make deep learning approaches such as convolutional neural networks impractical. Instead, we demonstrate that if a reasonable segmentation can be performed, specifically in the 3D space provided by CT scanning technology, then classification is fairly easy by standard classifiers. Our unsupervised segmentation technique is based on the scale-space decomposition algorithm known as the Sieve, which produces segmentations at different scales based on the location of extrema. The segmentation is represented as an unpacking of the bag into constituent elements. Each segment is characterised by a histogram of voxel intensities, which can then be classified. Segments at different scales may represent only part of an object (a battery inside a torch, for example). Hence, once the segments of a bag are classified, we map the predicted class labels back to the bag (effectively repacking the back) to provide a visualisation of the location of the devices. We call this system Unpack, eXtract, Predict, Repack (UXPR).

To assess this approach and compare to other algorithms we use the 3D volumetric CT database of scanned baggage made available by the ALERT {\em `Segmentation of objects from volumetric CT data'} initiative \cite{crawford2013segmentation}. This dataset was created using a medical CT scanner and contains ground-truth labeling of objects. The ground-truth labeling was performed in Mevislab (www.mevislab.de) which assists the user in drawing contours around an object. The data consists of 45 bags with a range of items, both electrical and non-electrical. Because of the limited amount of data, we also create simulated bags made up of samples of the constituent bags. All experiments follow a leave-one-bag-out methodology, i.e. we train on the segmentation of 44 bags and test on an unseen bag.

Our contributions can be summarised as follows: we show the utility of the segmentation approach by showing that if ground truth segmentation is known, then near perfect classification is possible (Section~\ref{experiment1}), even when the object has not been seen before (Section~\ref{experiment3}). An evaluation of UXPR system (Section~\ref{experiment2}) shows we can accurately detect devices when the ground truth segmentation is not known. Finally, we demonstrate the utility of 3D images by comparing performance on 2D images and highlight the difficulty of training deep learning algorithms with 3D images or small numbers of 2D images (Section~\ref{experiment4}).

Our conclusion is that our segmentation algorithm captures enough of the characteristics of devices to classify using simple intensity histograms. We show that standard classifiers perform well, but that one algorithm, the Hierarchical Vote Collective of Transformation-based Ensembles, HIVE-COTE~\cite{lines18hive}, is more accurate, has better sensitivity and produces better probability estimates.

Speed is obviously an important characteristic of any potential real time detection system. The segmentation algorithm is approximately linear time and the classification of new cases by HIVE-COTE is fast. Our approach offers the potential for linear time segmentation where new data can be added with rapid rebuilding of the classifiers. This means the corpus of known devices can easily be extended and shared. This is important since new devices and concealment techniques are constantly being developed.

This paper is structured as follows. Section~\ref{background} gives an overview of related research into detecting threats from luggage images. Section~\ref{dataset} describes the data in more detail, and Section~\ref{UXPR} describes the UXPR process. Section~\ref{results} provides the evidence to support our contributions and conclusions, which are summarised in Section~\ref{sec:conclusions}.

\section{Background}
\label{background}

\subsection{Classification of x-rays} The automatic detection of threats in luggage is an active research area, although the majority of work has focused on 2D x-ray images. Ba{\c{s}}tan {\em et al.}~\cite{bacstan2011visual} apply a Bag of Visual Words (BoVW) technique~\cite{csurka2004visual}, in combination with SIFT features~\cite{Lowe2004distinctive} and a Support Vector Machine to detect bags containing firearms. They conclude that analysis of x-ray imagery is challenging and using the straightforward implementation of BoVW does not perform well.

Kundegorski {\em et al.}~\cite{kundegorski2016using} further explore the use of various feature point detectors and descriptors within the BoVW approach. They examine combinations of, SIFT, SURF~\cite{Bay2008speeded}, FAST~\cite{Rosten2010faster}, ORB~\cite{Rublee2011orb}, KASE~\cite{Alcantarilla2012kaze}, DAISY~\cite{Tola2010daisy}, BRISK~\cite{Leutenegger2011brisk}, FREAK~\cite{Alahi2012freak} and AKAZE~\cite{Alcantarilla2013Fast} with a support vector machine (SVM) and random forest (RF) as the final classifiers. They found that that the best performance was achieved using a FAST-SURF feature detector and descriptor combination with a SVM, and overall that the SVM performed consistently better than RF for this task.

Over the last few years, convolutional neural networks (CNNs) have become a popular and strong alternative for image classification problems. Akcay {\em et. al}~\cite{akcay2018using} consider the use of CNNs with transfer learning within the context of detecting firearms in x-ray images and a six class classification problem (firearm, firearm-component, knives, ceramic-knives, cameras and laptops). They directly compare the classification performance of AlexNet~\cite{krizhevsky2012imagenet}, ResNet~\cite{he2016deep} and VGG~\cite{simonyan2014vgg} CNN networks, pretrained on the ImageNet database~\cite{russakovsky2015imagenet} against a BoVW approach with a SVM. They note that all the CNNs have superior performance over BoVW features (FAST-SURF) for firearm detection. However, for this task, best performance was achieved by training a SVM classifier on CNN features. Griffen {\em et. al}~\cite{griffin2019bagging} use features extracted from a pretrained CNN to detect appearance anomalies, of shape, texture and density in 2D x-ray images of luggage.

Megherbi {\em et. al}~\cite{megherbi2012comparison} use 3D shape descriptors to find bottles and firearms in 3D CT volumes. These descriptors require the object to first be segmented. They use a fuzzy connectedness segmentation algorithm~\cite{Udupa1996fuzzy}, which was previously shown to work best in~\cite{Wirjadi2007survey}, when compared to region growing, watershed, level sets and thresholding segmentation. Features are then generated from these segmented volumes using a combination of 3D Zernike descriptors and Histogram of Shape Index. These features are then classified using SVM, Artificial Neural Network, Decision Trees, Boosted Decision Trees and Random Forests. Results show that Histogram of Shape Index constantly outperforms 3D Zernike, but combined are at least the same or (in a few occasions) better. Best classification performance was achieved using SVM and Random Forests.

An alternative approach to 3D object detection is to use 3D interest points and descriptors. This approach mitigates the need to perform initial segmentation on the test data (although it is common that systems still need to be trained on segmented objects).  Flitton {\em et. al}~\cite{flitton2013comparison} first find interest points with 3D SIFT~\cite{Flitton2010object}, then compare the classification performance of five different features; simple density; density histograms; density gradient histograms; 3D-SIFT and 3D-RIFT. The conclusions are that the simpler density descriptors(DH and DHG) outperform the more complex ones (3D-SIFT, 3D-RIFT). This is attributed to low, anisotropic voxel resolutions and high levels of CT-artifacts.


\subsection{Image segmentation}
Our approach is to perform an unsupervised segmentation of the image prior to classification. We use the  Sieve operator~\cite{bangham1988}, to find areas of interest in an image. The sieve uses morphological scale-space operations, specifically openings and closings, or combinations of them, to filter an input signal by removing extrema of specific scales. It does this by applying flat structuring elements to an input signal, which unlike conventional morphological operators such as those used in granulometries, have a fixed size but variable shape. This ensures that the shape of the structuring element is not seen as artifacts in the simplified signal since the sieve is designed to filter extrema by size rather than a fixed shape. They were introduced as a one-dimensional non-linear scale-space decomposition algorithm in~\cite{bangham1996b, bangham1996ip, bangham1996a}, but can be extend to \emph{n}-dimensions by adopting techniques from graph morphology~\cite{Bangham96, bangham1996b, Bangham96b}.

\begin{figure}[!ht]
	\centering
		\includegraphics[width=.75\columnwidth]{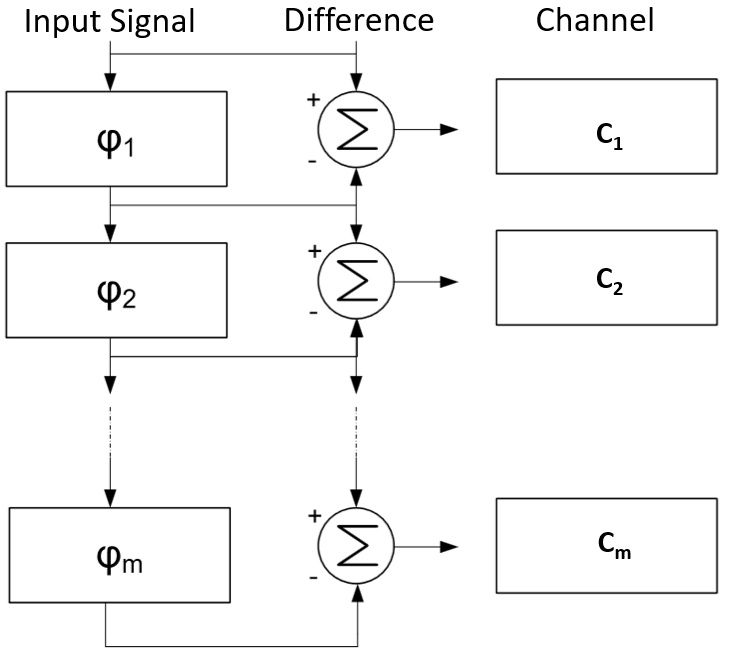}
	\caption{The structure of a sieve decomposition where $\varphi$ denotes the sieve operator. Channel images, $C$, can be formed by taking differences between each successive low-pass sieve output.}
	\label{fig:sieveStructure}
\end{figure}

In two-dimensions, as in an image for example, scale is a function of area, while in three-dimensions scale is a function of volume. At each stage the sieve operator $\varphi$ removes extrema of only that scale. At the first stage $\varphi_{1}$ removes extrema of scale 1, $\varphi_{2}$ removes extrema of scale 2 and so on until the maximum scale $m$. The maxium scale $m$ of a 1D is the signal is length, in 2D it is the number of pixels in the image and in 3D it is the number of voxels in the volume. Sieves preserve \emph{scale-space causality}~\cite{Bangham96} (as no new extrema are introduced at each stage) and can be described as a \emph{cascade} of morphological filters; as each stage in the filtering process is related to the previous. This serial structure, shown in Figure \ref{fig:sieveStructure}, can be contrasted with the parallel structure used in granulometries. Formally, the result of applying the sieve operator $\varphi$ at scale \emph{s} to an input signal \emph{f} can be written as,
\begin{eqnarray}
	f_s = \varphi_{s}(f_{s-1})
\end{eqnarray}
where the original signal (said to be at scale 0) is $f_0 \equiv \varphi_0(f) \equiv f$.

\begin{figure}[!ht]
	\centering
		\includegraphics[width=.45\textwidth]{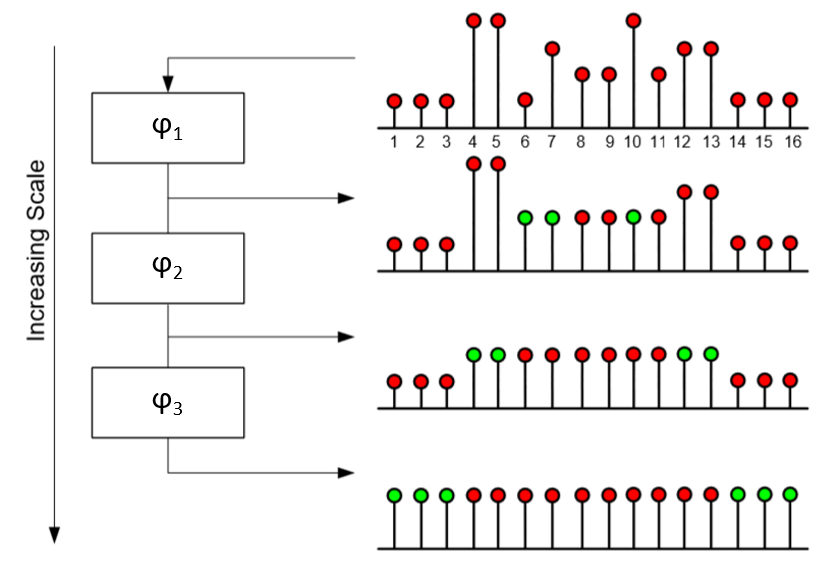}
	\caption{An example $\mathcal{M}$-sieve decomposition of a 1D signal. Green vertices are the vertices affected at each scale level.}
	\label{fig:Msievegraph}
\end{figure}

Figure~\ref{fig:Msievegraph} shows an example sieve decomposition of a 1-Dimensional signal. At the scale $\varphi_1$ the maxima and minima at points 6, 7 and 10 are smoothed to equal the nearest value of the neighbours. At scale $\varphi_2$ the extrema pairs at (4,5) and (12,13) are smoothed.

Figure \ref{fig:sieveStructure} also shows the \emph{Channel} domain of the sieve decomposition. The Channels, $C_s$, are the differences between each successive low-pass sieve output. These Channels are the non-zero elements of the channel functions and have the same position as they did in the original signal. Therefore the original input signal $f$ can be reconstructed from a simple summation of the Channels $C_s$ across all scales $s$ to the maximum scale $m$.

\begin{eqnarray}
	f = \sum^{m}_{s=1} C_s = C_1 + C_2 + C_3 \ldots + C_m
\end{eqnarray}
Because of this property the sieve can be said to be fully invertible and is a transform of the original signal.

The sieve operator $\varphi$ can be a morphological opening ($\psi$), closing ($\gamma$), $\mathcal{M}$-filter or $\mathcal{N}$-filters but not a dilation or erosion as they are not idempotent. In this report we opt to filter our signals using an $\mathcal{M}$-filter. An $M$-filter is defined as an opening filter followed by a closing filer and can be written as,
\begin{eqnarray}
	\mathcal{M}_rf(x) = \gamma_r(\psi_r(f(x))
\end{eqnarray}
If $\mathcal{M}_s$ is applied to a signal, this filter will first remove all maxima of scale $s$ and then minima of scale $s$.

\section{Dataset}
\label{dataset}
In this report we use a modified version of 3D volumetric CT database of scanned baggage made available by the ALERT {\em `Segmentation of objects from volumetric CT data'} initiative \cite{crawford2013segmentation}. This database was created using a medical CT scanner and the volumes were captured at a resolution of 512 x 512 x $\sim$512. We maintain the original XY volume dimensions but crop the volumes in the Z dimension to remove empty space and reduce overall volume size. In order to speed up computation, we also resample voxel intensity to be in the range 0 - 255. The database contains a total of 45 bags, packed with standard items normally found in traveller luggage. The 3D images have been manually segmented and each of 624 items has been labelled as one of 140 different types. We group the 140 class values in two ways: a two class non-electrical (543 cases)/electrical (81 cases) problem, and a five class problem:non-electrical, Mobile phone (12 cases), hard drive (8 cases) , laptop (4 cases), other electrical (57).

\begin{figure}
	\centering
		\includegraphics[width=0.4\columnwidth]{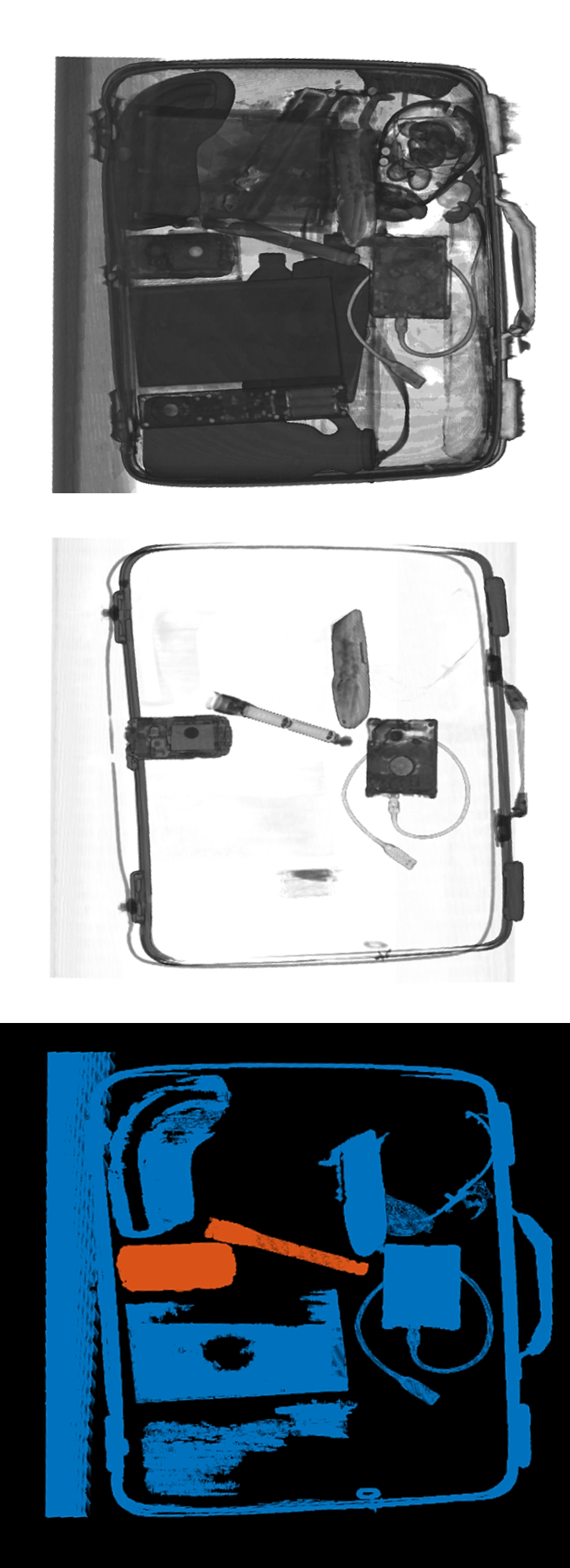}
	\caption{Top row shows a 3D CT scan of a bag (removing CT gantry and voxels of intensity less 28 to aid visualization). Middle row shows the resulting channel volume and connected sets of objects with voxel volumes greater than 20575 and less than 105830. Bottom row shows the result of classifying the channel image using 1-NN. Orange depicts that the object was classified as electronic and blue items are non-electronic. In this example the classifier correctly identifies the flashlight and mobile phone as electronics but miss-classifies the hard-drive as non-electric.}
	\label{fig:exampleBag}
\end{figure}

Example baggage CT-Volumes can be seen in Figure~\ref{fig:exampleBag} (top) and Figure~\ref{fig:repackClassify} (left column).

\begin{figure*}
	\centering
		\includegraphics[width=.75\textwidth]{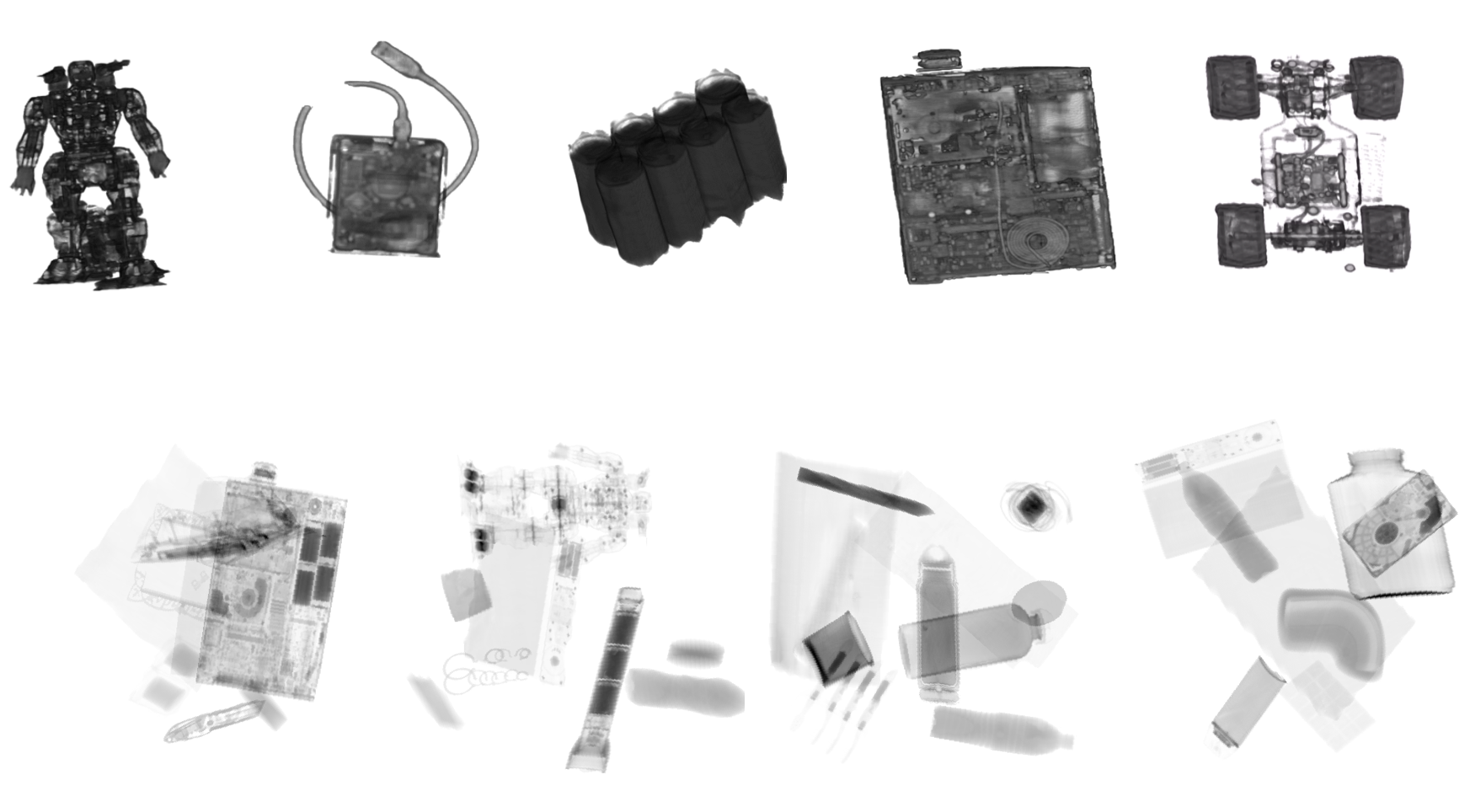}
	\caption{Top row shows example ground truth objects found in our bag database.These objects are then used to create a number of simulated bag volumes, examples of which are shown in the bottom row.}
	\label{fig:simulatedVolumes}
\end{figure*}

Because of the limited amount of volumes available, we create simulated bags from the constituents of the real bags. We do this by extracting every ground-truthed object from every bag to form a master pool of objects. Example objects are shown in Figure~\ref{fig:simulatedVolumes} (top row). Then a 'packing' algorithm randomly selects 20 objects from this master-pool, randomly rotates and then randomly places each object within a 512 x 512 x 512 volume. Each object is allowed to touch but no part of the object is allowed to intersect with another object or lay outside the 512 x 512 x 512 volume. Five attempts are made at placing each object into the volume, after which the object is abandoned and the algorithm moves onto the next object. This results in a simulated bag volume containing a variable number of objects, akin to what is found in real baggage. We generate 100 random simulated bag volumes which we use in our experiments in section~\ref{results}. Examples of these volumes are shown in Figure~\ref{fig:simulatedVolumes} (bottom row).

For all experiments we perform a leave one bag out experiment, i.e. we train on the contents of 44 bags and test on the contents of one.

\section{Unpack, eXtract, Predict and Repack (UXPR)}
\label{UXPR}
\begin{figure}
	\centering
		\includegraphics[width=0.9\columnwidth]{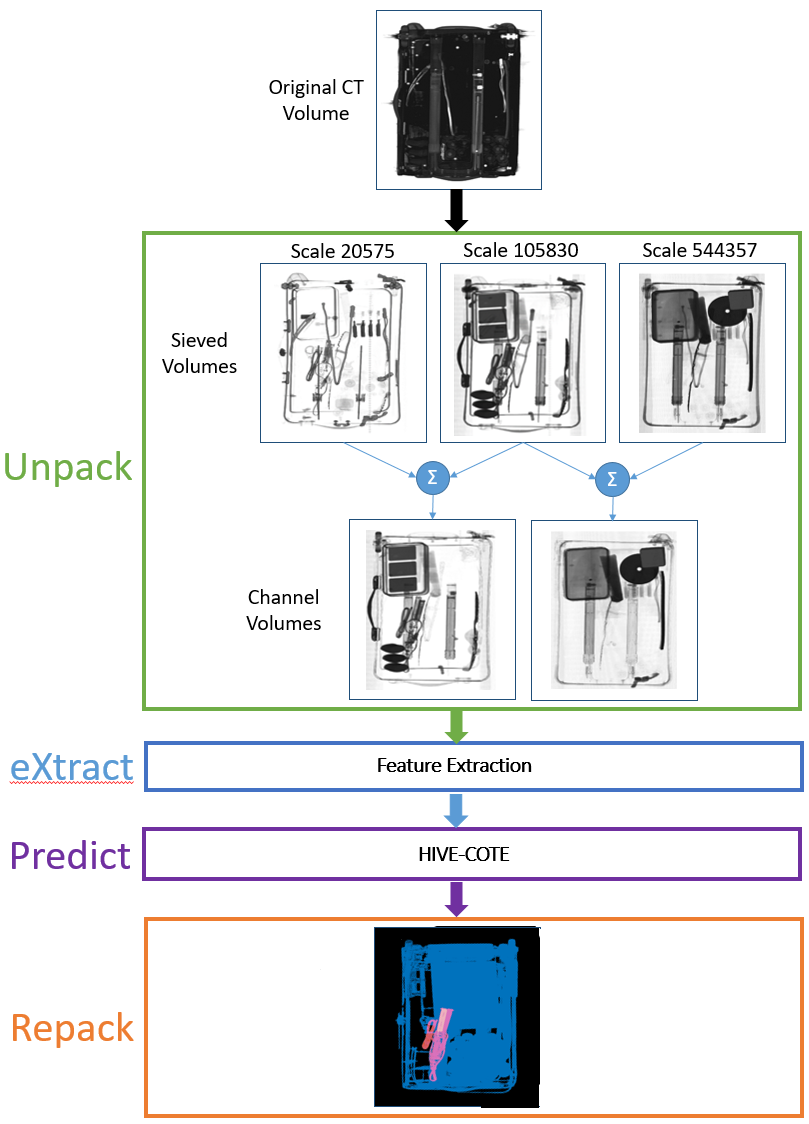}
	\caption{The original 3D CT scan is $Unpacked$ using the sieve algorithm described in section \ref{sec:Unpack}. In the $eXtract$ stage, histogram density features are then derived from objects in the Channel Volumes. These features are then classified in the $Predict$ stage using HIVE-COTE. Bags are then $Repacked$ by recombining classification labels and colour coded depending on predicted class.}
	\label{fig:uxpr}
\end{figure}

We adopt a piecewise approach to the detection of electric devices,  summarised in Figure \ref{fig:uxpr}. This involves:
\begin{enumerate}
\item Unpacking the bag using the Sieve unsupervised segmentation algorithm;
\item eXtracting features from each item detected in the bag;
\item Predict probability estimates of whether each segment is an electric device or not;
\item Repacking the back to summarise the number and location of devices.
\end{enumerate}

\subsection{Unpack}
\label{sec:Unpack}
The first step is to `unpack' the bag using the Sieve algorithm described in section~\ref{background}. We Sieve each CT-volume to $N$ different scales, $\left[S_{1}\ldots S_{N}\right]$ where $log_{10}S_{n}$ are equispaced between 0 and $log_{10}P$, where $N=5$ and $P=2800000$ were chosen to remove all objects from the CT-volume. This results in five sieved volumes of scale, 4000, 20575, 105830, 544357 and 2800000 with each volume containing a varying number of connected-sets of voxels. Channel volumes, $C_{n}$, are then formed by subtracting each sieved volume from one sieved to a previous scale, $C_{n}=\left|\varphi_{n}-\varphi_{n-1}\right|$ \ ; \ $n = \left[2\cdots N\right]$. Each of these four Channel volumes will contain connected-sets of voxels which can either correspond to a complete segmented object, or part of an object or a collection of connected objects.

\subsection{eXtract}
We extract features from each of the connected-sets in each Channel volume. We use simple density histograms for features (0 to 255) since it was shown in~\cite{flitton2013comparison} that simpler histogram density descriptors achieved better classification performance in 3D CT imagery, when compared to more complex interest point detectors such as SIFT and RIFT. Furthermore, histograms have the benefit of generating a feature vector of consistent length, irrespective of the volume of the object and are invariant to the rotation and position of the object. The histograms are not normalised for size as the volume of the objects they represent is an important feature.

\subsection{Predict}
 A histogram is a vector of real valued attributes, hence any classifier could be used. Our choice of classifiers is guided by three factors: we wish to find the most accurate and sensitive approach; we desire good probability estimates to help improve decision making; and we want to explore whether there is discriminatory information in the ordering of the attributes.

 For a baseline, we perform whole image classification using a deep learning configuration based on previous experimentation~\cite{akcay2018using}. For classifying histograms we use random forest~\cite{breiman01randomforest}, rotation forest~\cite{rodriguez06rotf} and XGBoost~\cite{chen16xgboost}. These are widely used and have claim to represent state of the art~\cite{bagnall18rotf}. We also use a meta ensemble of all three, the heterogenous ensemble of standard classifiers (HESCA)~\cite{large17hesca}. HESCA uses a probability weighted voting scheme for constituent classifiers, and has been shown to improve the probability estimates of the base classifiers. We have experimented with tuned support vector machines, but the performance was poor. We omit the results in the interest of clarity.

For detecting more complex discriminatory features in the histogram, we use time series classification (TSC) algorithms~\cite{bagnall17bakeoff}. TSC problems involves classifying any ordered real valued series (i.e. not necessarily ordered in time). TSC algorithms are potentially useful in scenarios where there may be some warping between attributes (i.e. the intensity calibration between bags may change), when phase independent shapes are relevant (e.g. the number of peaks or shape of a specific peak are discriminatory) or when autocorrelation between features defines class (e.g. co-location of two peak a certain distance apart defines class). We use the Hierarchical Vote Collective of Transformation-based Ensembles, HIVE-COTE~\cite{lines18hive} to detect features in these domains. HIVE-COTE combines ensembles using five different representations: an ensemble of nearest neighbour classifiers using elastic distance measures, the Elastic Ensemble~\cite{lines15elastic}; a bag of words approach ensembled over different parameter values, BOSS~\cite{schafer15boss}; a frequency domain classifier Random Interval Spectral Ensemble~\cite{lines18hive}; an interval based method using summary statistics, Time Series Forest~\cite{deng13forest}; and the Shapelet Transform~\cite{bostrom17binary} which finds phase independent discriminatory subseries. The probabilistic predictions of the classifier built on each representation are combined with the HESCA ensemble method. HIVE-COTE has been shown to outperform other TSC approaches on a range of benchmark problems~\cite{bagnall17bakeoff}.

\subsection{Repack}
The aim of this stage is to produce a visualization that will aid security CT-operators in their screening procedures. After each channel-object in each channel volume has been classified, the final stage is to repack the items into a single volume highlighting the position of detected electronic devices.  Repacking is done by recombining the classification label of each of the channel volumes on a voxel to voxel basis. We use a simple voting system to designate a voxel as either $very-unlikely-electronics$ or $unlikely-electronics$ or $likely-electronics$. If a voxel is classified as electronics in two or more channel-volumes then we designate that voxel $likely-electronics$. If the voxel is classified as electronics on only one channel-volume then it is designated as $unlikely-electronics$. If the voxel is never classified as electronics then it is designated as $very-unlikely-electronics$.

\section{Results}
\label{results}

\subsection{Why not deep learning?}
\label{results:cnn}

Working with CNNs on the raw 3D volumes is very difficult. This is largely due to the massive input size relative to, especially in our case, the number of example bags we have access to learn from. If we consider the raw 3D volumes of 512x512x$\sim$512 voxels, the input layer for the network would be $\approx$134 million nodes, which would then be passed through the various network layers. Even the largest of 2D networks (see Table \ref{tab:tferlearners}) have around this many parameters in total, which are learned on extremely large datasets. Nevertheless, we attempted a modest network architecture to determine feasibility of a direct 3D-volume to item classification procedure. This comprised of a network with two convolutional layers, a fully connected layer, and a final softmax prediction layer, for classifying segmented item volumes. Item volumes were scaled and if necessary padded to 32x32x32 while maintaining aspect ratio for input into the networks, to standardise and allow for a reasonable training time. We experimented with fixed hyperparameters and a grid search over a space of 256 parameter sets. In both instances, the resulting classifier produced predictions that were no better than picking the majority class. We conclude that training a 3D network is both resource intensive and not effective, at least with a small number of images.

Working in 2D gives us more options, since there has been significant work in this area with impressive results. In particular, a transfer learning approach~\cite{oquab2014learning} becomes possible, as considered in \cite{akcay2018using} on a firearm detection problem. 
Transfer learning refers to using a large, complex network that has had its weights trained on a different (much larger) dataset (usually ImageNet \cite{russakovsky2015imagenet}), which is then used on a different dataset with the final prediction layers removed. A new softmax layer may be trained, or the outputs of the intermediate layers extracted and used as features in a different classifier.

The expectation is that the network has learned `generically good' image classification features, because it has trained on such a large number of cases from thousands of different classes. The aim is to make use of these features and re-purpose them for the problem at hand. Because of this, we can overcome - at least to a large degree - the lack of data problem when considering neural networks for this task. Also, working in 2D, we need to downsample the images far less to achieve manageable input spaces. Finally, training time for a deployable system is vastly reduced if pretrained weights can be leveraged, as opposed to training from random initialisation. In the field, thousands of new images would be generated daily which can feed back in the training of new models, especially as new, unseen electrical devices appear in the future. Therefore a system that does not require weeks to train once a situation where plentiful training data is arrived at is desirable.

Table \ref{tab:tferlearners} summarises the networks considered in this study. All implementations are available through the Keras API~\cite{chollet2015keras}, including the pretrained ImageNet weights.

\begin{table}[htb]
    \centering
       \caption{Summaries of the pretrained CNNs considered, all from \texttt{https://keras.io/applications/}. Acc refers to the Top-1 accuracy of the network on the ImageNet validation dataset.}
    \begin{tabular}{c|ccc}
        \hline
         Network & \#Paras & Depth & Acc \\ 
         \hline
         InceptionResNetv2~\cite{szegedy2017inception} & 55M & 572 & 0.803 \\             
         Inceptionv3~\cite{szegedy2016rethinking} & 23M & 159 & 0.779 \\         
         ResNet50~\cite{he2016deep} & 25M & 168 & 0.749 \\                   
         VGG16~\cite{simonyan2014vgg} & 138M & 23 & 0.713 \\           
         VGG19~\cite{simonyan2014vgg} & 143M & 26 & 0.713 \\             
         Xception~\cite{chollet2017xception} & 22M & 126 & 0.790 \\             
         \hline
    \end{tabular}
    \label{tab:tferlearners}
\end{table}

We evaluate whether the transfer learning method can detect the presence of `threats' given a 2D image of a full bag. For this formulation of the dataset, `threat' refers to a mobile phone or laptop being present in the bag. Flattening the original scans into 2D images of course discards depth information, and makes segmentation far more difficult. Tightly packed bags will inevitably result in occlusion of deeper objects of some form, no matter how the transform takes place. We therefore test the viability of going directly from images to threat classification via the network-generated features, as opposed to including a segmentation step and use of hand-engineered features. The dataset used in this transfer learning sub-study therefore is comprised of 45 2D images of bags. 15 bags contains mobile phones or laptops, and are thus labelled as threats. We continue to evaluate using leave-one-bag-out cross validation.

The transferal process may involve fine tuning, where the pretrained network is also trained on the new data of interest, to give the network the chance to adapt to any markedly different properties in the new dataset. Our dataset indeed has a different distribution to that found in the majority of ImageNet: it is greyscale (the intensities are duplicated across RGB channels for input into the networks); and is generally less complex than many pictures taken in a natural-environment context.

We performed preliminary experiments to determine the efficacy of fine tuning the networks for this dataset with relatively conservative parameters. We searched the number of end-layers to tune, with the remaining layers being frozen, $L = \{1 \dots 5\}$, and the number of epochs, $E = \{ 5, 10, 20, 50 \}$, using the Adam optimiser \cite{kingma2014adam}. These spaces are conservative for two reasons. First, the lack of data, even with the benefit of transfer learning requiring less training examples. Secondly, more optimisation means more computation, thus decreasing the benefit of training speed provided by transfer learning. We found no consistent benefit or trend in performance over this search space, and therefore conclude that given the amount of data available to us at this time, extensive searching for improvements would not be a reliable use of resources. The remainder of this section considers transfer learning with no fine tuning.

\begin{figure}[!ht]
		\includegraphics[width=1.\linewidth, trim={0cm 1cm 1.5cm 0cm},clip]{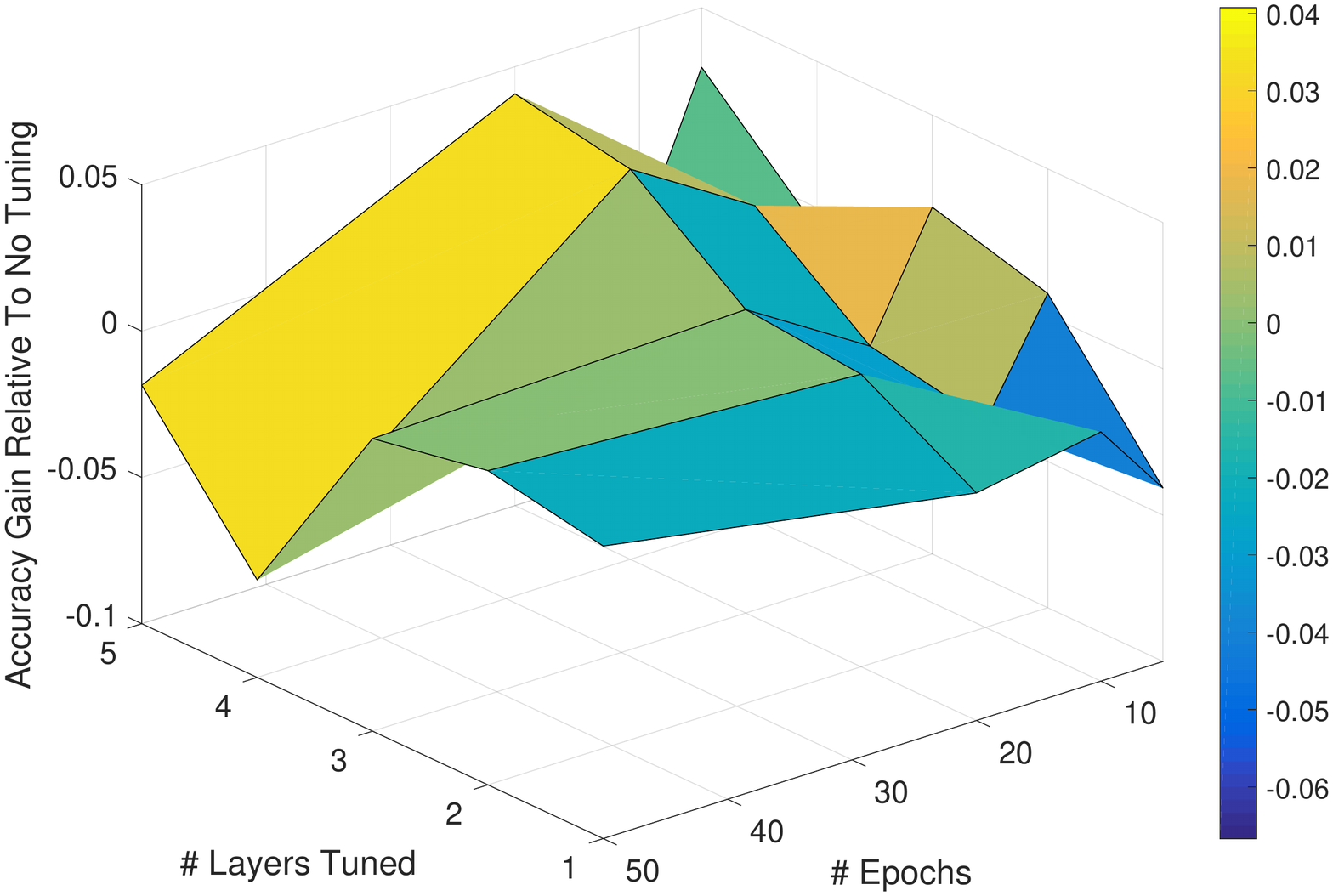}
	\caption{Averaged over the six networks considered, the accuracies of different fine tuning parameters expressed as the difference between them and if we performed no fine tuning at all. No consistent benefit is found.}
	\label{fig:finetuning}
\end{figure}

Figure \ref{fig:finetuning} shows results averaged over the six pretrained networks, with 1-NN as the final classifier (which proved to be best on average, see Table \ref{tab:coreTransferLearningResults}), as the surface of accuracy gain relative to the networks with no fine tuning. For the parameters considered, no consistent improvement in accuracy or potential trends were found. We conclude that given the amount of data available to us at this time, extensive searching for improvements would not be a reliable use of resources. The remainder of this section considers transfer learning with no fine tuning.

Once the network-generated features are extracted, we evaluate a range of classifiers for completeness.  Table~\ref{tab:coreTransferLearningResults} summarises the performance of seven classifiers on each of the pretrained networks. Accuracy is given across the 45 bags, with the number of threats missed in parentheses out of a total of fifteen. The classifiers are: one nearest neighbour with Euclidean distance (1-NN); support vector machine with quadratic kernel (SVM); rotation forest (RotF); extreme gradient boosting (XGBoost); heterogeneous ensemble of standard classifiers (HESCA); and random forest (RandF).	

There are three main conclusions to be drawn from this. First, a simple nearest neighbour is surprisingly effective, achieving the highest average and individual accuracy. The forests of trees were, on average, only as good as predicting the majority class (0.66). The outputs of the base networks are high dimensional (1028 up to 4096) and continuous, and therefore 44 instances are perhaps not enough for them to learn from while meaningfully diversifying their trees. Second, the best pretrained networks for this particular problem appear to be InceptionResNetv2 and Xception. In fact, the relative performances of each network aligns closely with the reported Top-1 accuracies on the original ImageNet dataset in Table \ref{tab:tferlearners}, with the exception of ResNet50, which appears to have had a harder time transferring to the new problem. Lastly, these results show that for this particular problem with the data we currently have, leveraging transfer learning and network generated features cannot immediately solve the problem of threat detection in bags. The single best result was using ED on Inceptionv3, where a total of four errors were made: three threats missed, and one non-threat misclassified as a threat. However, we do not have enough results to make claims of significance that it is these two particular methods in combination that are the best.

Perhaps of more interest is the only case where no threats are missed: using an SVM and InceptionResNetv2. In general, the SVM achieved the highest sensitivity of all classifiers regardless of network. InceptionResNetv2 (IResNetv2 in the Table) was tied with Xception as being the most sensitive model. Given that an SVM can easily be altered to further optimise against different costs of misclassification \cite{masnadi2012cost} such as those obviously present in this problem, further work in this direction along with the production of larger quantities of data should continue to consider SVMs, as well as nearest neighbour classifiers.

\begin{table*}[!ht]
    \centering
        \caption{Accuracies (and number of threats missed, of fifteen), of each classifier on the threat or not threat dataset transformed by each of the six pretrained networks.}
    \begin{tabular}{c|ccccccc}
        \hline
        Base Network	 & 1-NN	 & SVM	 & RotF	 & XGBoost	 & HESCA	 & RandF	& Mean \\
        \hline
        InceptionResNetv2	 & 0.778 (5)	 & 0.889 (0)	 & 0.756 (8)	 & 0.756 (8)	 & 0.756 (9)	 & 0.622 (14)	 & 0.759 \\
        Inceptionv3	 & 0.911 (3)	 & 0.711 (8)	 & 0.711 (10)	 & 0.733 (10)	 & 0.733 (11)	 & 0.644 (15)	 & 0.741 \\
        ResNet50	 & 0.711 (8)	 & 0.644 (4)	 & 0.489 (15)	 & 0.578 (13)	 & 0.578 (15)	 & 0.6 (15)	 & 0.6 \\
        VGG16	 & 0.778 (5)	 & 0.756 (2)	 & 0.622 (12)	 & 0.644 (13)	 & 0.667 (12)	 & 0.622 (14)	 & 0.682 \\
        VGG19	 & 0.822 (3)	 & 0.733 (4)	 & 0.711 (10)	 & 0.667 (9)	 & 0.667 (11)	 & 0.644 (14)	 & 0.707 \\
        Xception	 & 0.889 (4)	 & 0.8 (5)	 & 0.778 (7)	 & 0.711 (9)	 & 0.733 (8)	 & 0.733 (10)	 & 0.774 \\
        Average	 & 0.815	 & 0.756	 & 0.678	 & 0.682	 & 0.689	 & 0.644	 & 0.710 \\
        \hline
    \end{tabular}%
    \label{tab:coreTransferLearningResults}
\end{table*}


\subsection{If we have the correct segmentation, can we accurately detect electric devices?}
\label{experiment1}

\begin{figure}[!ht]
		\includegraphics[width=\linewidth, trim={1cm 9cm 0cm 9cm}]{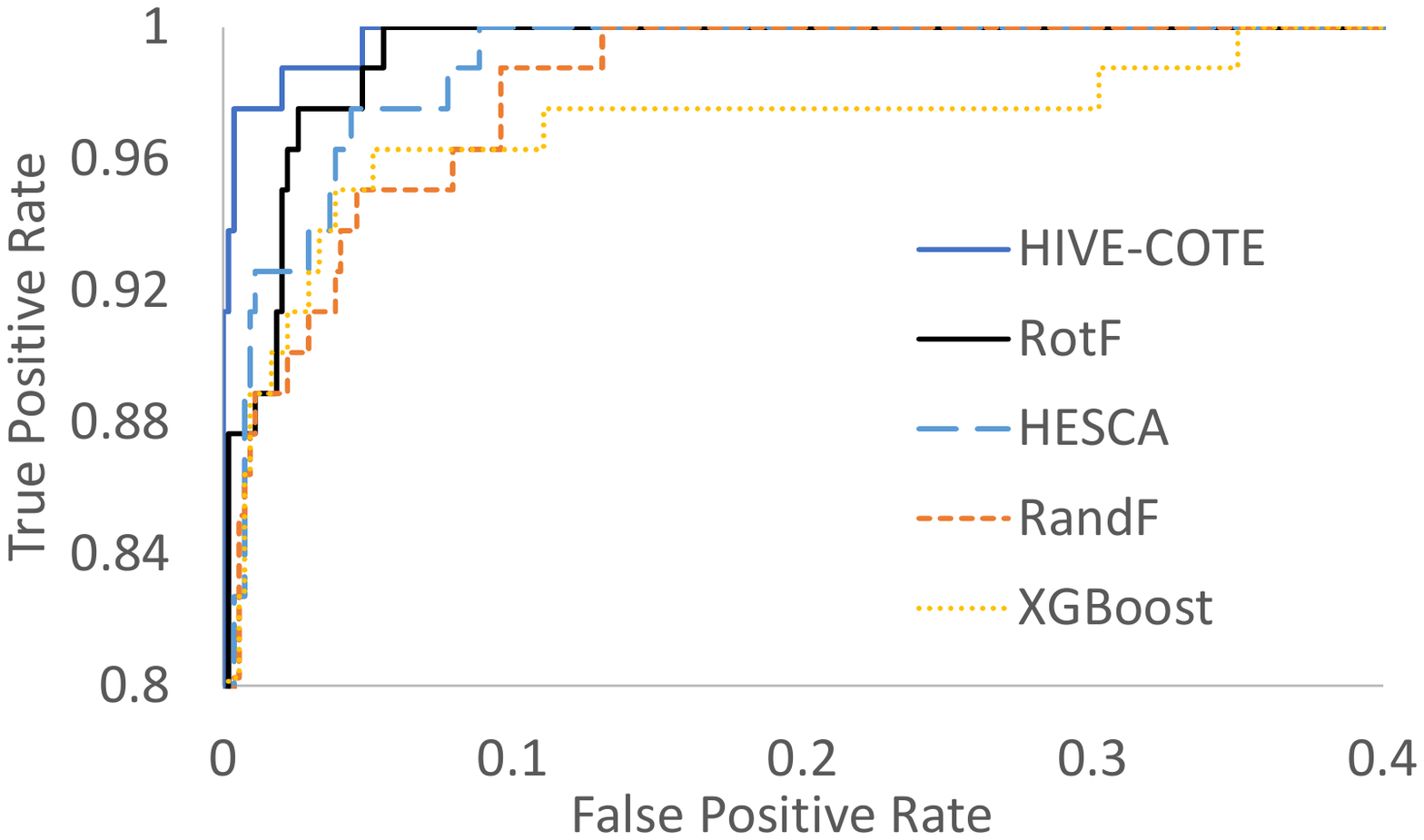}
	\caption{Receiver operator characteristic (ROC) curves for five classifiers on the two class problem where the ground truth segmentation is known.}
	\label{fig:rocExperiment1}
\end{figure}

The first set of segmentation experiments examines how accurately we can classify items given the correct segmentation of a 3D image. These experiments serve as a basic sanity check of the suitability of the features used and an indication of classifier selection. Table~\ref{tab:GTtoGT} presents the accuracy and number of mis-classified cases for five classifiers on four problems: both original and simulated two class and five class problems.
\begin{table*}[!ht]
    \centering
       \caption{Leave one bag out accuracy and number mis-classified cases for five classifiers on four problems where the ground truth segmentation is known. }
    \begin{tabular}{c|cccc}
Classifier   & Two Class & Two Class Simulated & Five Class & Five Class Simulated \\ \hline
1-NN                & 98.56 (9) & 98.02 (7)     & 98.08 (12) & 98.28 (7)\\
Random Forest       & 97.60 (15) & 99.15 (3)    & 96.96 (19) & 98.77 (5) \\
Rotation Forest     & 98.08 (12) & 98.87 (4)  &   97.11 (18)    & 97.79 (9)\\
XGBoost             & 97.43 (16) &  99.15 (3)    & 97.43 (16)  & 99.02 (4) \\
HESCA               & 97.43 (16) & 98.87 (4)    & 96.96 (19) &  98.53 (6) \\
HIVE-COTE           & 98.88 (7)  & 99.71 (1)    & 98.40 (10)  & 99.26 (3) \\
\hline
    \end{tabular}
    \label{tab:GTtoGT}
\end{table*}
As we expected, 1-NN performs as well as any other time domain classifier when the ground truth segmentation is known. This result supports our base hypothesis: if we can segment well, a complex deep learning approach is not required. Nearest neighbours are slow to classify new instances and do not produce good probability estimates. Fast estimation of probabilities will ultimately be essential for informed risk assessment. It is also worth noting that HIVE-COTE is the best performing algorithm on all four data sets. Given the high overall accuracy of all classifiers and the fact there are a different number of cases in each bag, we cannot meaningfully test for significant differences over folds. However, the HIVE-COTE results strongly indicate that there is discriminatory information in the interaction of the intensity values, not just in the values themselves.  Table~\ref{tab:twoClass} shows the area under the receiver operator curve (AUROC) and the negative log likelihood (NLL) for five classifiers and Figure~\ref{fig:rocExperiment1} shows the associated ROC curves (zoomed into the top left corner for clarity). Higher AUC and lower NLL are preferable. HIVE-COTE has the highest AUC and the lowest NLL, followed by rotation forest, then HESCA.

\begin{table*}
    \centering
    \caption{Area under the receiver operator curve (AUC), the negative log likelihood (NLL), sensitivity and specificity for six classifiers on the two class problem where the ground truth segmentation is known. }
    \begin{tabular}{c|cccc}
Classifier   & AUROC & NLL  & Sensitivity & Specificity\\ \hline
1-NN                & N/A    & N/A  & 0.9383    & 0.9926 \\
Random Forest       & 0.9913  & 75.69  & 0.8519 & 0.9945\\
Rotation Forest     & 0.9963 & 59.13  & 0.8642 & 0.9982 \\
XGBoost             & 0.9862 & 81.71  & 0.8519 & 0.9926  \\
HESCA               & 0.9947 & 61.41 & 0.8519 & 0.9926\\
HIVE-COTE           & 0.9989 & 42.76 & 0.9382 & 0.9963  \\         \hline
    \end{tabular}
    \label{tab:twoClass}
\end{table*}

There are certain items that are mis-classified by all classifiers, and this is due to dubious labelling or poor image quality/segmentation. For example, one item that all classifiers get wrong is a phone in a case, where the case is included in the segmentation but the internals of the phone are not. These discrepancies can confound the classifier, suggesting that an alternative approach using partial segmentations or alternative classifiers may yield benefits.

\subsection{Can we accurately detect electric devices with an unsupervised segmentation?}
\label{experiment2}
\begin{figure}[!ht]
		\includegraphics[width=\linewidth, trim={1cm 9cm 0cm 9cm}]{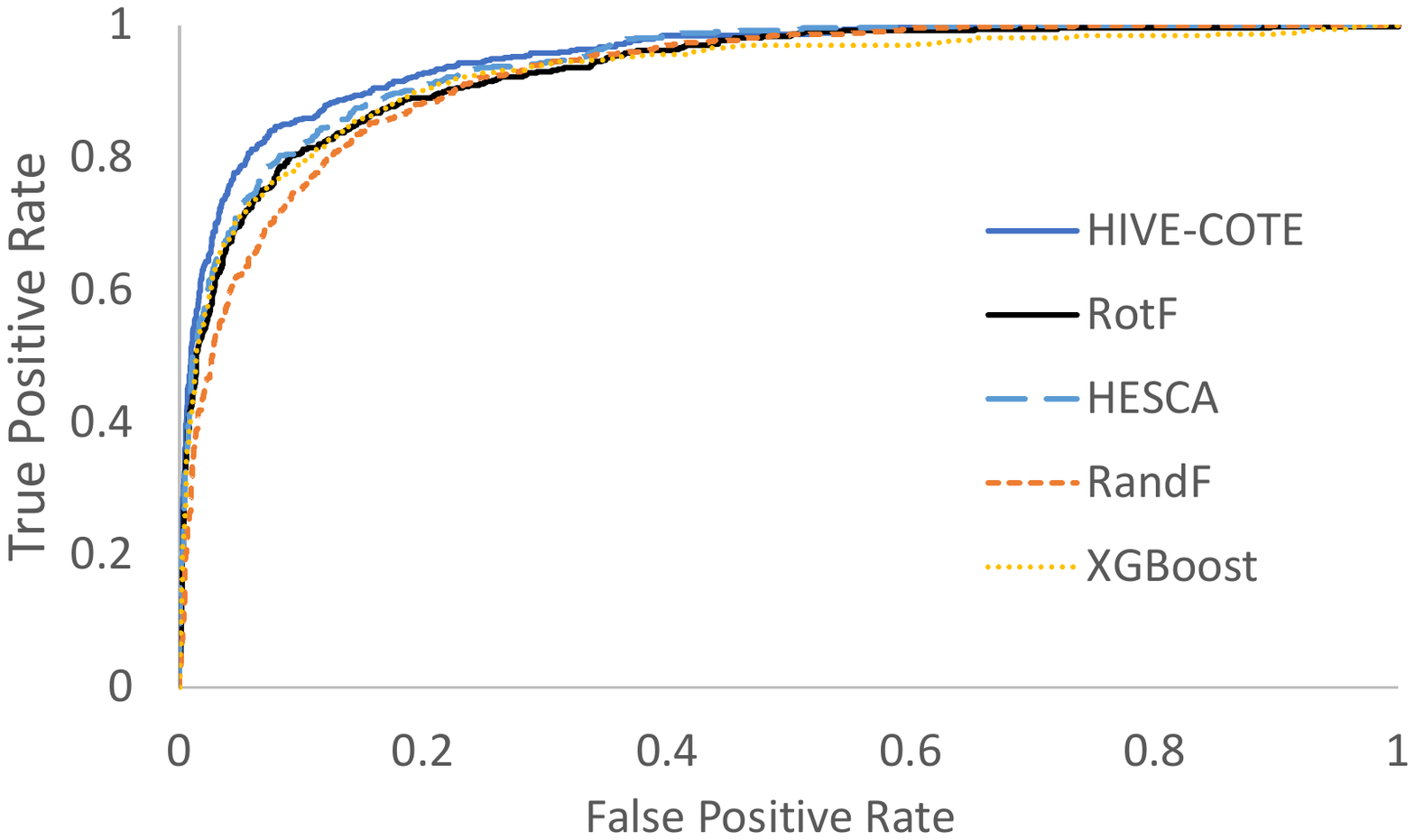}
	\caption{ROC curves for five classifiers on the two class problem where the ground truth segmentation is not known.}
	\label{fig:rocExperiment2}
\end{figure}

\begin{figure}[!ht]
		\includegraphics[width=\linewidth]{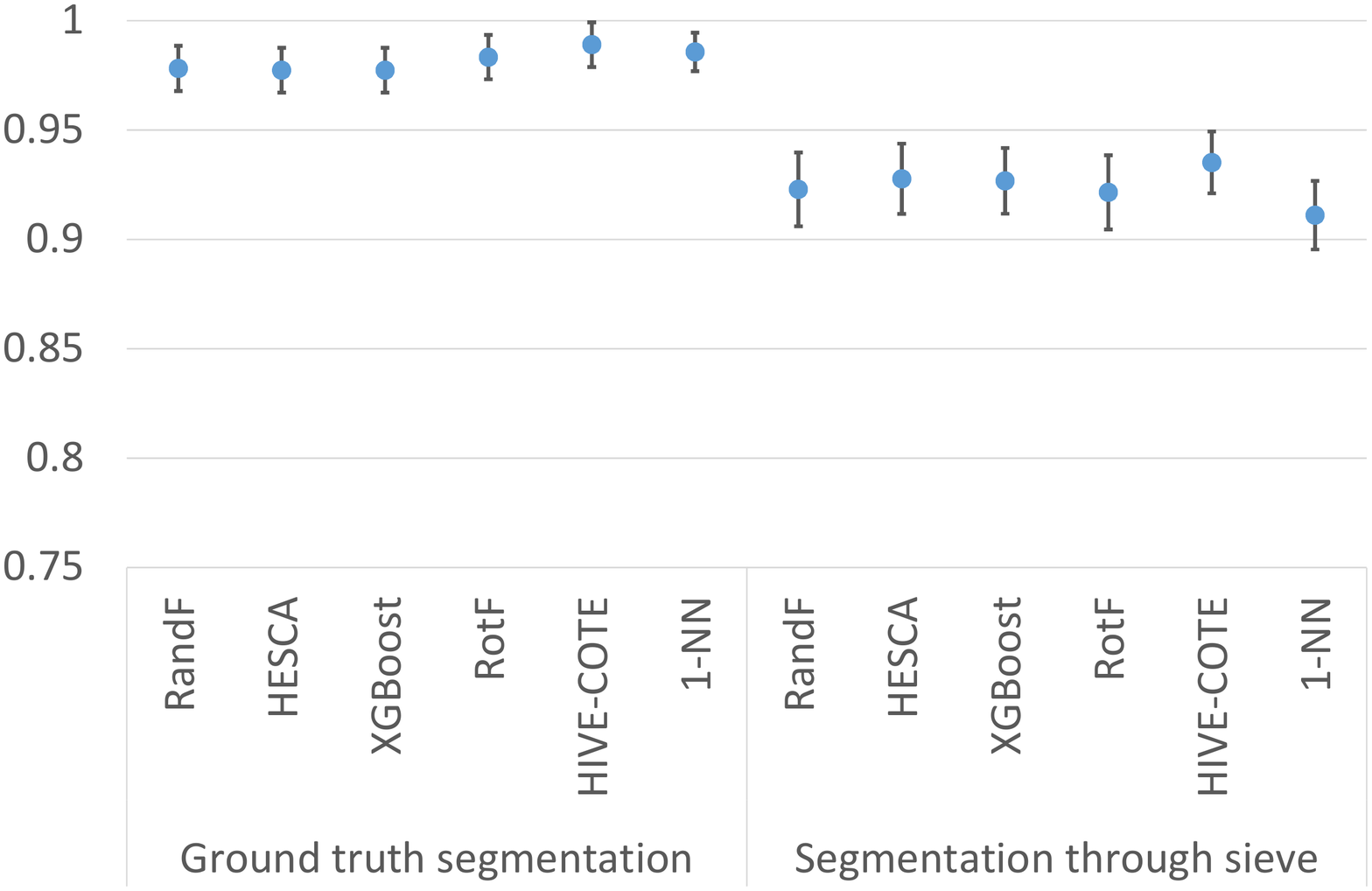}
	\caption{Comparison of performances between classification on manually (left side) and automatically (right side) segmented items as being electrical or not. The data are averaged over the 45 bags and the bars represent the 95\% confidence interval.}
	\label{fig:gtsegVSseiveseg}
\end{figure}

\begin{figure*}[!htb]
	\centering
		\includegraphics[width=0.75\textwidth]{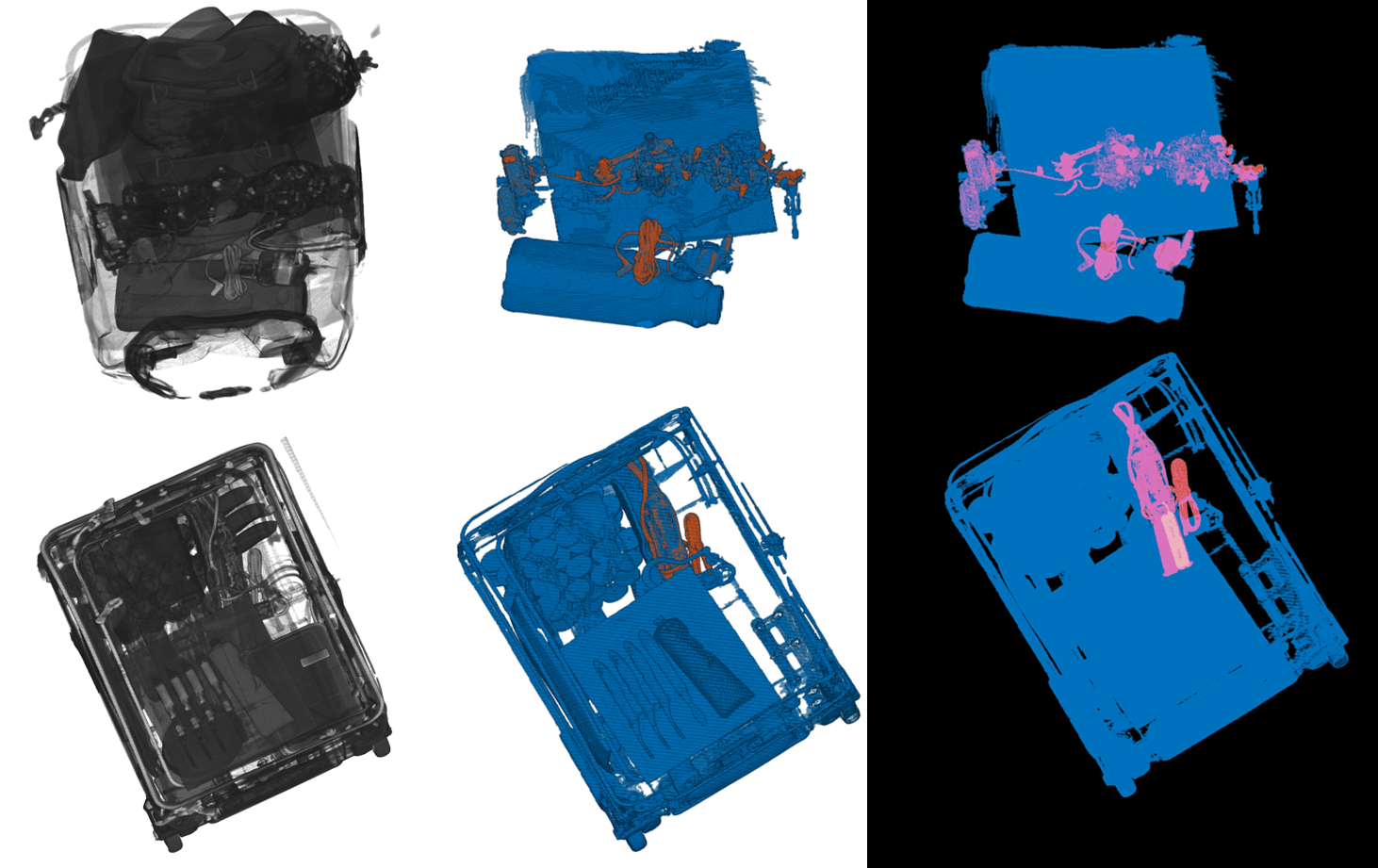}
	\caption{Two examples of CT baggage scans (left column). CT scans have been thresholded to only show voxels with intensity greater than 27 to aid visualization. For each bag the middle column shows surface renderings of the results from the repacking step. Objects that are $unlikely-electronics$ are shown in blue and items that are $likely-electronics$ are shown in orange. Objects that are classified $very-unlikely-electronics$ are not shown. The right most column shows maximum intensity projections through each voxel to aid object visualization and location in bag.}
	\label{fig:repackClassify}
\end{figure*}

The UXPR system uses a fast segmentation algorithm that may not necessarily extract whole items. Rather, it extracts segments at different scales. To investigate whether we can still classify devices using the unsupervised sieve, we perform the segmentation on each bag and manually label whether each segmentation is a constituent part of an electrical device. We label the segmentation as electrical if it overlaps in any way with the device. Table~\ref{tab:unsupres} shows the performance statistics for six classifiers on the unsupervised problem. Figure~\ref{fig:rocExperiment2} shows the ROC curves for the five classifiers that produce a probabilistic output.
\begin{table*}[htb]
    \centering
        \caption{Classification accuracy for six classifiers with unsupervised segmentation.}
    \begin{tabular}{c|ccccc}
Classifier          & Accuracy & AUROC & NLL & Sensitivity & Specificity\\ \hline
1-NN                &  0.9042  &  NA   &  NA       & 0.6698 &  0.9426 \\
Random Forest       &  0.9076  & 0.9235  &   1244    & 0.5151     & 0.9719 \\
Rotation Forest     &  0.9185  & 0.9315  &   1147    & 0.5773 & 0.9743 \\
XGBoost             & 0.9228   & 0.9257  &   1318    & 0.6226 & 0.9719 \\
HESCA               & 0.9244   & 0.9426  &   1065    & 0.6094 & 0.9759 \\
HIVE-COTE           & 0.9307   & 0.9502  &   1013    & 0.6698 & 0.9734 \\ \hline
    \end{tabular}
    \label{tab:unsupres}
\end{table*}
HIVE-COTE has the highest accuracy, AUROC and NLL, and, perhaps of most significance, the highest sensitivity. The ROC curve shown in Figure~\ref{fig:rocExperiment2} highlights the superiority of HIVE-COTE. HESCA performs marginally better than the other standard classifiers.
A comparison of accuracy for the supervised and unsupervised segmentation is shown in Figure~\ref{fig:gtsegVSseiveseg}.
The most significant decrease in accuracy of 1-NN confirmed our prior belief that a more complex algorithm would be required for unsupervised classification. The lower accuracy and in particular the lower sensitivity (see Table~\ref{tab:unsupres}) does not mean we are not finding devices. Each segmented case may contain only a small portion of the device, and cases at different scales overlap. The repacking of the bag by overlaying the predictions at different sieve representations is more indicative of the detection process.
Figure \ref{fig:repackClassify} shows examples from the Repacking stage with $unlikely-electronics$ shown in blue and $Likely-electronics$ shown in orange. Voxels that are $very-unlikely-electronics$ are not shown. The back-pack in Figure \ref{fig:repackClassify} (top row) contains an electronic toy-robot and an AC adapter and both these objects are correctly identified as $likely-electronics$. Finally the hair straighteners (with power cord) are correctly identified as the only $Likely-electronics$ device in the suitcase, Figure \ref{fig:repackClassify} (bottom row).

\subsection{Can we accurately detect electric devices not seen before?}
\label{experiment3}
We have demonstrated that if an electric item is in the training data we can accurately identify it using the UXPR system. However, the range of devices available in practice and the constant stream of new devices being launched make this assumption unrealistic. To simulate the problem of an unseen device we construct a third set of experiments involving leaving one of the eighteen classes of electric device out of the training data, and testing on a simulated bag containing the unseen device. For the test data we insert the unseen electric devices along with a number of non electrical devices at the same proportion as the original data. Table~\ref{tab:unseen} presents the results for five classifiers. We use the ground truth segmentation for these experiments, so the results are comparable to those presented in section~\ref{experiment1}. Again, HIVE-COTE is the best performing algorithm, with a decrease in accuracy of just 1.5\%. Sensitivity of 77.78\% indicates that generalisation to unseen devices is possible in many cases, given a good segmentation.

\begin{table*}[htb]
\centering
    \caption{Classification accuracy for six classifiers on the leave one device out problem.}
    \begin{tabular}{c|ccccc}
Classifier          & Accuracy & AUROC & NLL & Sensitivity & Specificity\\ \hline
Random Forest       & 0.9668   & 0.9730  &   124   & 0.7037 & 0.9982\\
Rotation Forest     & 0.9643  &  0.9775 &   111   &  0.6790 & 0.9972 \\
XGBoost             & 0.9630  & 0.9734 &  150    & 0.6790 & 0.9957 \\
HESCA               & 0.9464  & 0.9591  & 165     & 0.5061 &  0.9972 \\
HIVE-COTE           & 0.9745  & 0.9956  & 85     & 0.7778 & 0.9972 \\ \hline
    \end{tabular}
    \label{tab:unseen}
\end{table*}
\begin{figure}[!ht]
		\includegraphics[width=\linewidth, trim={1cm 9cm 0cm 9cm}]{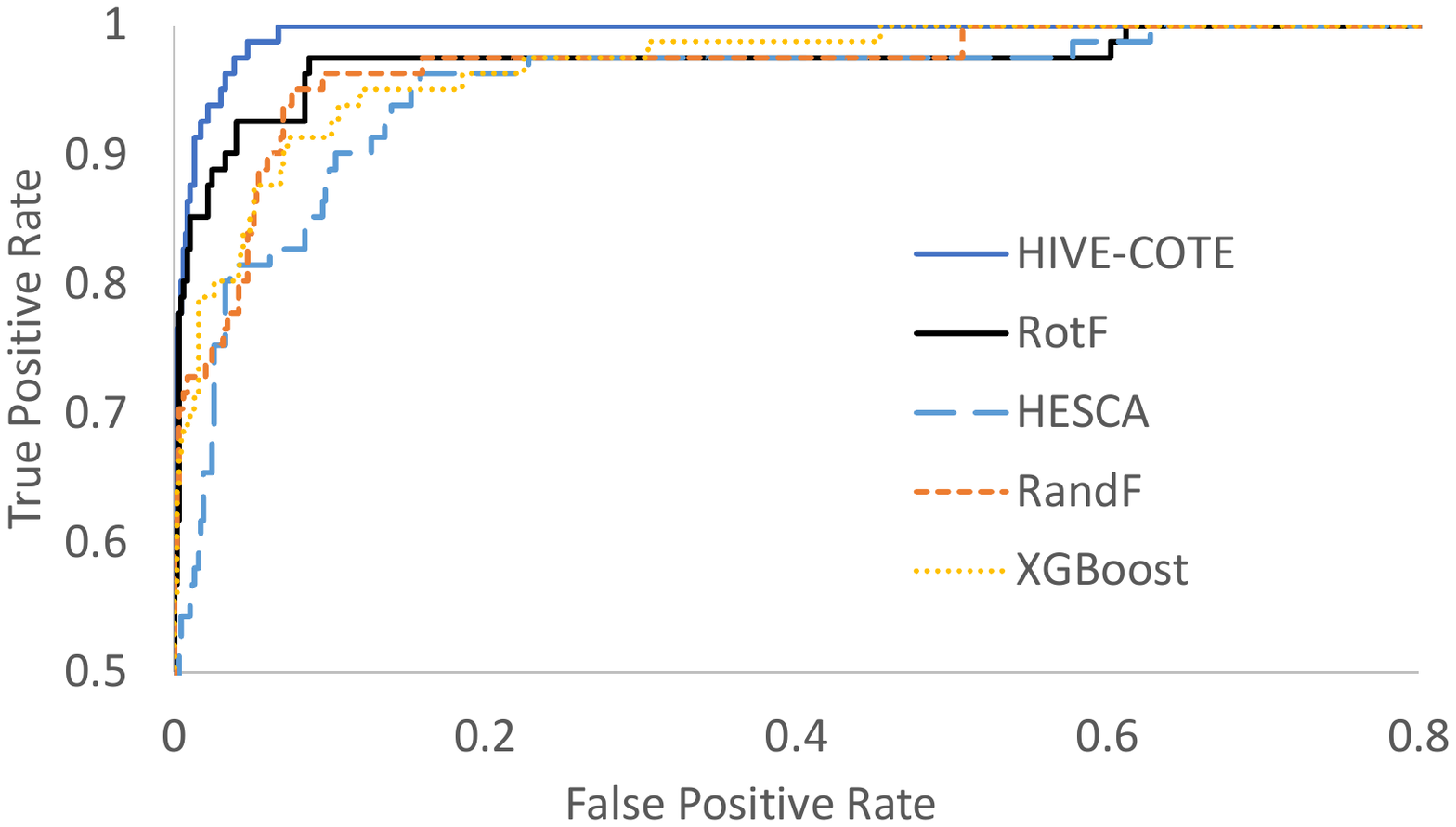}
	\caption{Receiver operator curves for five classifiers where the device in the test set is not in the train set (using manual segmentation).}
	\label{fig:rocExperiment3}
\end{figure}

\subsection{Is classifying 3D images easier than classifying 2D?}
\label{experiment4}
\begin{figure}[!ht]
\centering
		\includegraphics[width=.38\textwidth]{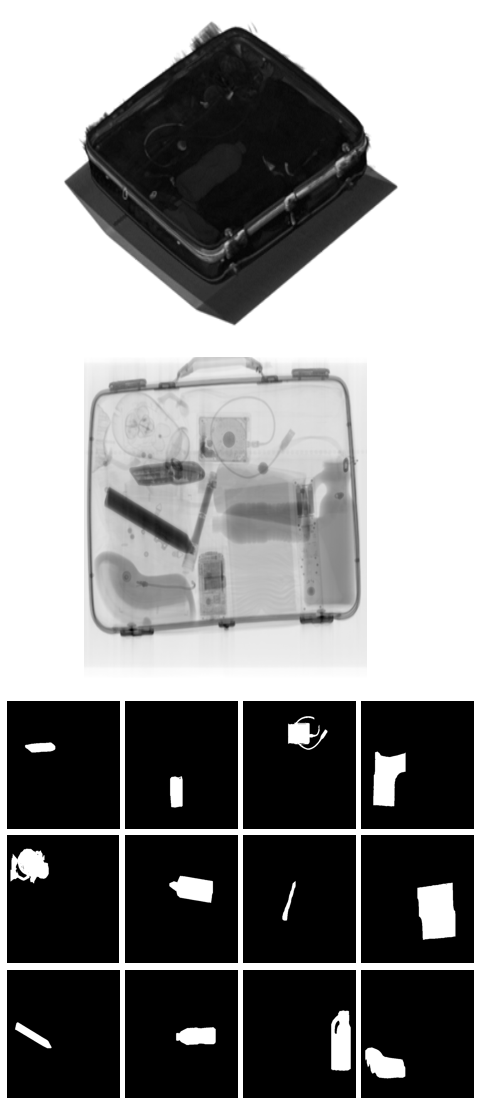}
	\caption{Figure shows the original 3D volumetric CT scan (top row) flattened down into a 2D projection (middle row). New 2D ground truth labels were also created (bottom row) by projecting along the same axis.  }
	\label{fig:2Dflat}
\end{figure}

3D scanners are more expensive than 2D and represent a significant investment in equipment and retraining of staff. Standard whole image classification algorithms such as CNN are impractical in 3D because of the high dimensionality and the commensurate long train times and difficulty in parameterisation. We investigate whether the move from 2D to 3D is actually worthwhile. We flatten the 3D image of each bag into 2D images by summing all voxels in the 3D scan along a chosen axis and then normalizing by the maximum values. Corresponding ground truth labels were also generated by summing the volumetric ground truth volumes along the same axis. Examples of the projected 2D image and ground truth are shown in Figure~\ref{fig:2Dflat}.
\begin{table*}[htb]
    \centering
        \caption{Classification accuracy for six classifiers on the 2D classification problem.}
    \begin{tabular}{c|ccccc}
Classifier          & Accuracy & AUROC & NLL & Sensitivity & Specificity\\ \hline
1-NN                & 0.8205  & NA &  NA  & 0.5682 & 0.8619 \\
Random Forest       & 0.8638  &  0.8067 & 297   & 0.1111 & 0.9761 \\
Rotation Forest     & 0.8621  & 0.8062 & 295    & 0.1111 & 0.9742 \\
XGBoost             & 0.8702  &  0.7977  & 416   & 0.2469 & 0.9632 \\
HESCA               & 0.8686  & 0.8128  & 292   & 0.1728 & 0.9724\\
HIVE-COTE           & 0.9038 & 0.8830  & 239 &  0.4198 & 0.9761 \\ \hline
    \end{tabular}
    \label{tab:2D}
\end{table*}
\begin{figure}[!ht]
		\includegraphics[width=\linewidth, trim={1cm 9cm 0cm 9cm}]{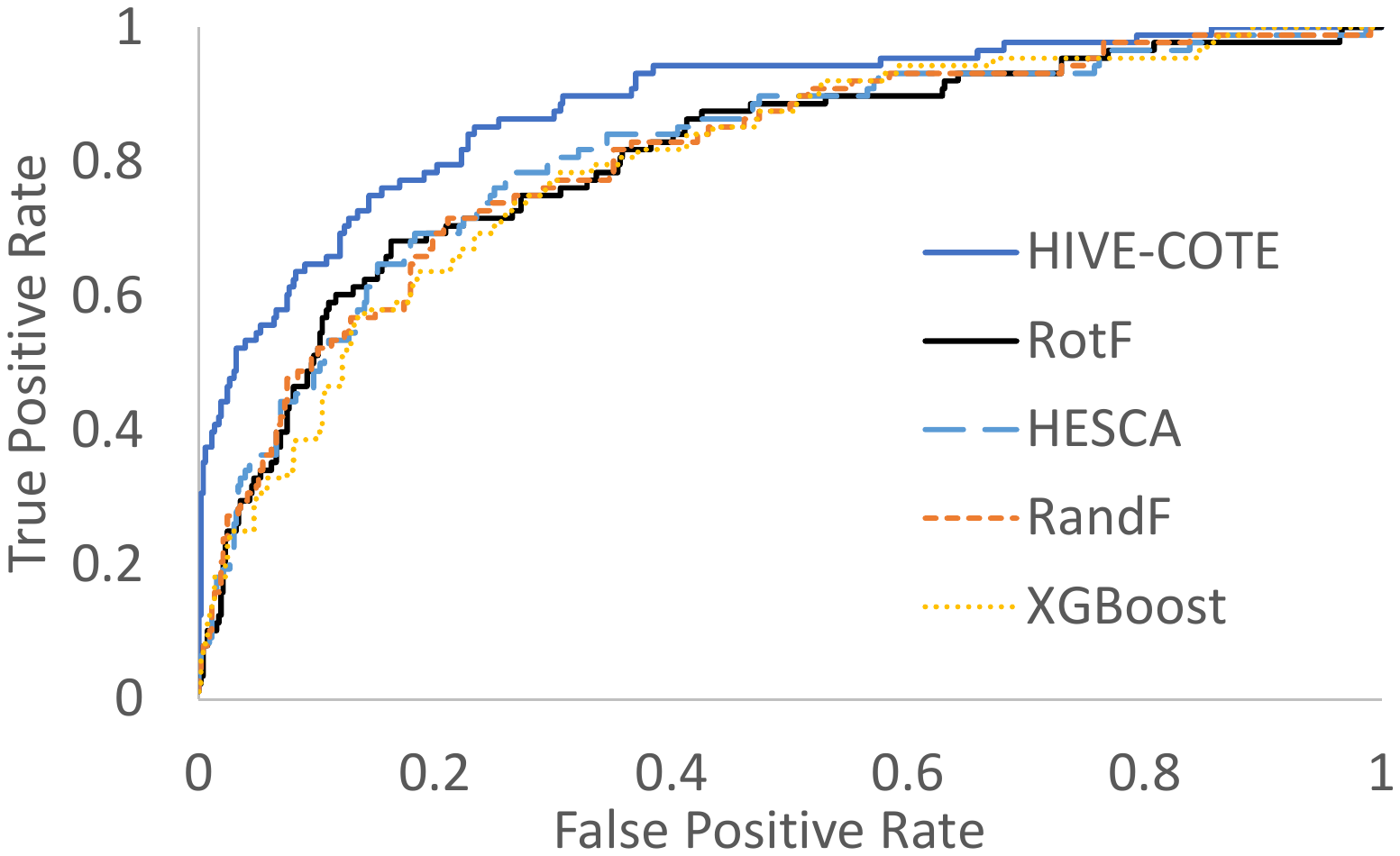}
	\caption{ROC curves for five classifiers using the 2D images.}
	\label{fig:rocExperiment4}
\end{figure}

\begin{figure}[!ht]
		\includegraphics[width=.95\linewidth]{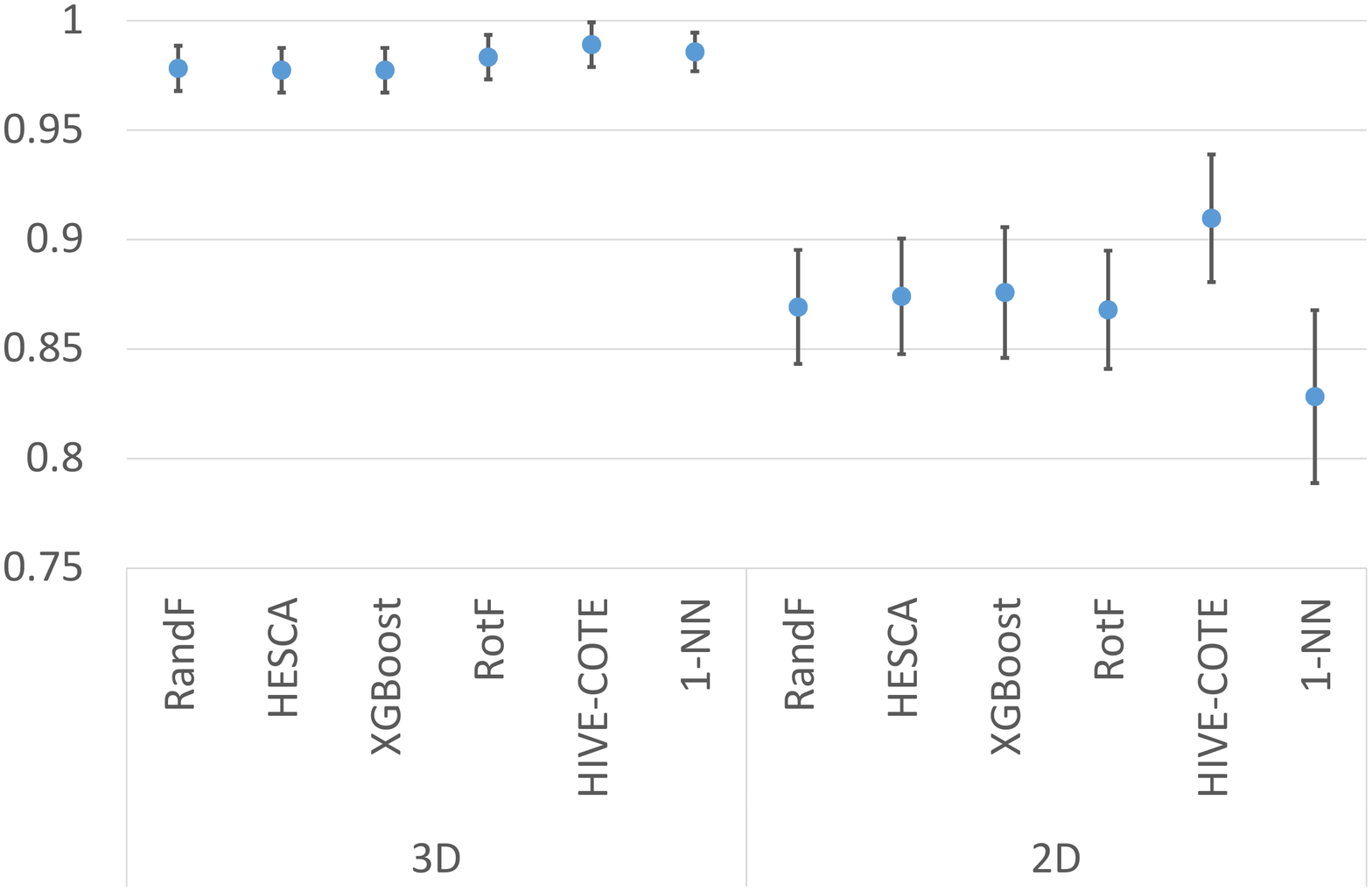}
	\caption{Comparison of performances between classification of items being electrical or not when considering the 3D volumes (left side) and 2D flattened images (right side).}
	\label{fig:3dVS2d}
\end{figure}

Table~\ref{tab:2D} shows the classification results with the 2D images. The associated ROC curves are shown in Figure~\ref{fig:rocExperiment4}. Figure~\ref{fig:3dVS2d} shows the performance change when moving from 3D to 2D. All the classifiers are significantly less accurate using 2D, most notably in sensitivity. All the classifiers except HIVE-COTE are doing little better than predicting the majority class. HIVE-COTE drops in sensitivity from 93\% to 42\%. This supports our hypothesis that classifying electric devices is significantly easier with 3D images. In the long term, the fact that automation is more practical with 3D images may justify the start up costs involved in deployment.


\section{Conclusions}
\label{sec:conclusions}

3D scanners are more expensive than 2D machines and their images can be harder to manually manipulate and interpret. Nevertheless, the greater amount of information available in a 3D scan makes automated processing more feasible. We have described a system for detecting electric devices in 3D scans of luggage that is based on unpacking the bag through an unsupervised segmentation, predicting whether segments are electrical or not based on intensity histograms, then repacking the bag through overlaying segmentations taken at different scales. We have tested the feasibility of this system with a dataset generated for the DHS ALERT program and have shown that we can achieve a high level of accuracy, robustness and interpretability. We have demonstrated that for the data we have, UXPR works well. The next stage of this research is to gather a greater amount of data. This will allow us to explore whether more complex methods for each stage improve the overall performance. Unpack currently just extracts segmentations at four fixed resolutions but using a more adaptive, data driven method of determining the resolution and spatially linking segments may provide more information for classification. Currently, eXtract involves forming the voxel intensity histogram for the whole segment. There are many more features we could extract that also may yield a more accurate system in the predict stage. Repack could make use of probabilistic predictions and better recombination schemes derived from the linkage of segments. These improvements will be assessed in a process of gradually moving to more realistic scenarios, with the ultimate goal of deploying a prototype integrated with a scanner tested in the real world.

\section*{Acknowledgements}
We would like to thank Northeasten University and the ALERT {\em `Segmentation of objects from volumetric CT data'} initiative \cite{crawford2013segmentation} for allowing us to use their CT-baggage database. Part of this work was supported by the UK Engineering and Physical Sciences Research Council (EPSRC)  [grant number EP/M015807/1]. The experiments were carried out on the High Performance Computing Cluster supported by the Research and Specialist Computing Support service at the University of East Anglia and using a Titan X Pascal donated by the NVIDIA Corporation.

\end{document}